\documentclass{article}

\usepackage{arxiv}

\usepackage[utf8]{inputenc} 
\usepackage[T1]{fontenc}    
\usepackage{hyperref}       
\usepackage{url}            
\usepackage{booktabs}       
\usepackage{amsfonts}       
\usepackage{nicefrac}       
\usepackage{microtype}      
\usepackage{cleveref}       
\usepackage{lipsum}         
\usepackage{graphicx}
\usepackage{natbib}
\usepackage{doi}
\usepackage{CJKutf8}
\usepackage{graphicx}

\title{ArthModel: Enhance Arithmetic Skills to Large Language Model}

\date{}

\author{ {\hspace{1mm} Yingdi Guo} \\
	\texttt{eteced@gmail.com}\thanks{Can be touched via this email also 2016101877@alu.ruc.edu.cn.} \\
}

\renewcommand{\headeright}{Draft}
\renewcommand{\undertitle}{Draft}
\renewcommand{\shorttitle}{ArthModel: Enhance Arithmetic Skills to Large Language Model}

\hypersetup{
pdftitle={ArthModel: Enhance Arithmetic Skills to Large Language Model},
pdfsubject={ai, llm},
pdfauthor={Yingdi Guo},
pdfkeywords={First keyword, Second keyword, More},
}

\begin{document}
\maketitle

\begin{abstract}
With the great success of ChatGPT, the research of large language models has become increasingly popular. However, the models have several limitations, such as toxicity and pool performance of arithmetic solving. Meanwhile, LLM may have some potential abilities that have yet to be exploited. In this paper, we choose a different way to enhance the arithmetic ability of LLM. We propose to train LLM to generate a postfix expression related to the arithmetic problem and incorporate it with small pretrained models. Moreover, this small model transfers the token embeddings into real dense numbers and invokes native functions of a deep learning platform to get the correct answer. To generate the final result, we propose prompt injection for adding the result outputs by the small model to LLM. This work provides different ways of thinking, training and using a language model. The codes and models will be released at \url{https://github.com/eteced/arithmetic_finetuning_v1}.
\end{abstract}

\keywords{ArthModel \and LLM}

\section{Introduction}
\begin{CJK}{UTF8}{gbsn}
There are works armed for extending the ability of LLM. By adding the chain of thought for an associated answer, \cite{wei2022chain} improves the ability of large language models to perform complex reasoning. \cite{xu2023wizardlm} rewrites the prompts to upgrade the simple instruction to a more complex one and increase diversity for LLM training. \cite{schick2023toolformer} trained the model to use different API calls and get the correct answers. \cite{gao2023pal} do more, they incorporate LLM with a python interpreter that the LLM generates codes for the python interpreter and gets the answer from it, which improves the accuracy of the answer. However, to make LLM solve arithmetic problems well, there could be a more straightforward way to do it. More specifically, the model should be able to access the native calls of a deep learning platform it runs on. Making LLM cooperate with a small model can also be thought as LLM incorporating a small cheap API.

In this paper, we take LLaMA as an example and illustrate how LLM incorporates a small pretrained arithmetic-solving model to enhance its arithmetic ability. We first train the LLM model to generate a postfix expression for an arithmetic calculation problem, which can be thought of as a 'code' input for the small arithmetic-solving model and a CoT for the problem since it determined which numbers and math ops should be considered first. Then, the small arithmetic-solving model will turn the token vectors into real dense numbers and invoke native function calls like 'add' to perform the math operations on dense numbers. The deep learning platform provides native functions that also support the running of LLM. Finally, incorporating the result from the small arithmetic-solving model, the LLM outputs the desired answers.

In the rest of this paper, we will first illustrate the framework of the work and then dig into the details of the small arithmetic-solving model and the methods to train such a model. Next, we show how to train LLM to output right for the small models. Finally, we use the Alpace data with the arithmetic data we generated to train the ensembled model. With the prompt injection, LLM can generate the correct arithmetic result without losing the general chat ability.
\end{CJK}

\section{Framework}

\begin{figure}
    \centering
    \includegraphics[width=0.8\columnwidth]{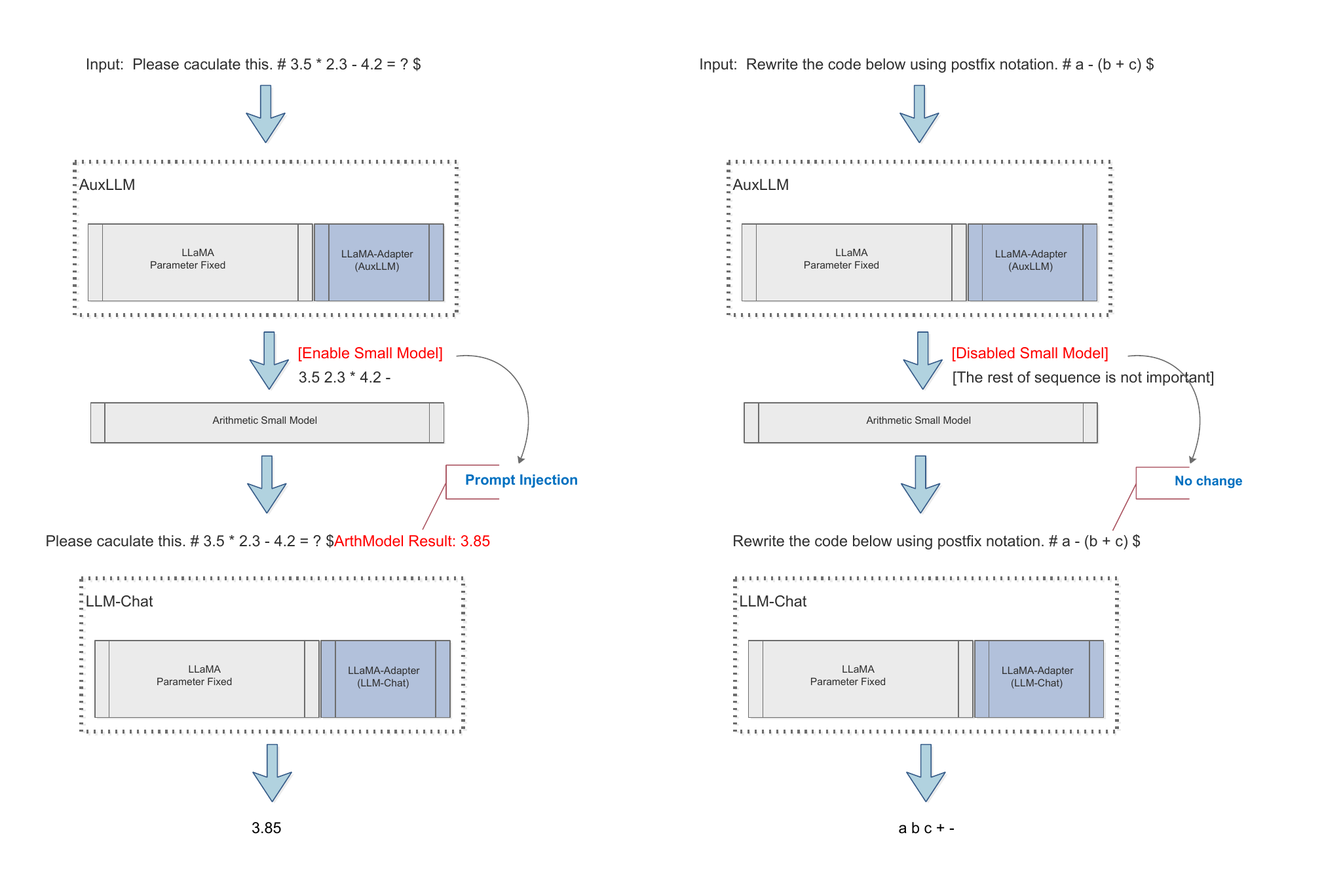}
    \caption{Overview of the framework. The left side shows the workflow when input is an arithmetic question, while the right side shows the workflow when input is a general question.}
	\label{fig:fig1}
\end{figure}

As shown in Figure \ref{fig:fig1}, there are three models in our framework. The AuxLLM model is trained by LLaMA-Adapter, which classifies whether the input is an arithmetic question and outputs the postfix expressions. An arithmetic small model called ArthModel stably turns the postfix expression tokens into dense numbers and math ops,  then calculates the math result. The LLM-Chat model is the final answer-generating model that gets the inputs from both the ordinate input and the output of ArthModel if the ordinate input is an arithmetic question. Note that AuxLLM model and LLM-Chat model share the same fixed LLaMA backbone while using different LLaMA-Adapter parameters.

\label{sec:framework}

\subsection{AuxLLM Model for Postfix Expression Prediction}
The primary purpose of the AuxLLM model is to generate a sequence that determines whether enable the arithmetic small model and the rest of the sequence to feed the small model.
Given the ordinate tokens input sequence $t_1, t_2, ..., t_n$, the AuxLLM model outputs the sequence  $o_1, o_2, o_3, ..., o_n$ that is the input for the arithmetic small model, we named ArthModel. Especially, the first character $o_1$ indicates whether to enable ArthModel and do prompt injection, as the embedding of $o_1$ is the input of the activated gate.

Unlike the regular LLM models, we append extra fixed-length white spaces at the ends of the ordinate input sequence before appending the tokens of correct answers at training progress. Since the extra white spaces are part of the input, the loss function will ignore the generation at those positions and that could be considered as an extra draft area for AuxLLM as we do not require the first token of the correct answer should be given once the last token of ordinate input is given. Moreover, we consider some tokens to be much more important to generate than others. More specifically, in our arithmetic problem-solving case, the decimal dot and math operations should have less tolerance for a mistaken prediction.  In this case, we add extra loss weight in such token prediction.

\subsection{ArthModel}
\begin{figure}
    \centering
    \includegraphics[width=0.4\columnwidth]{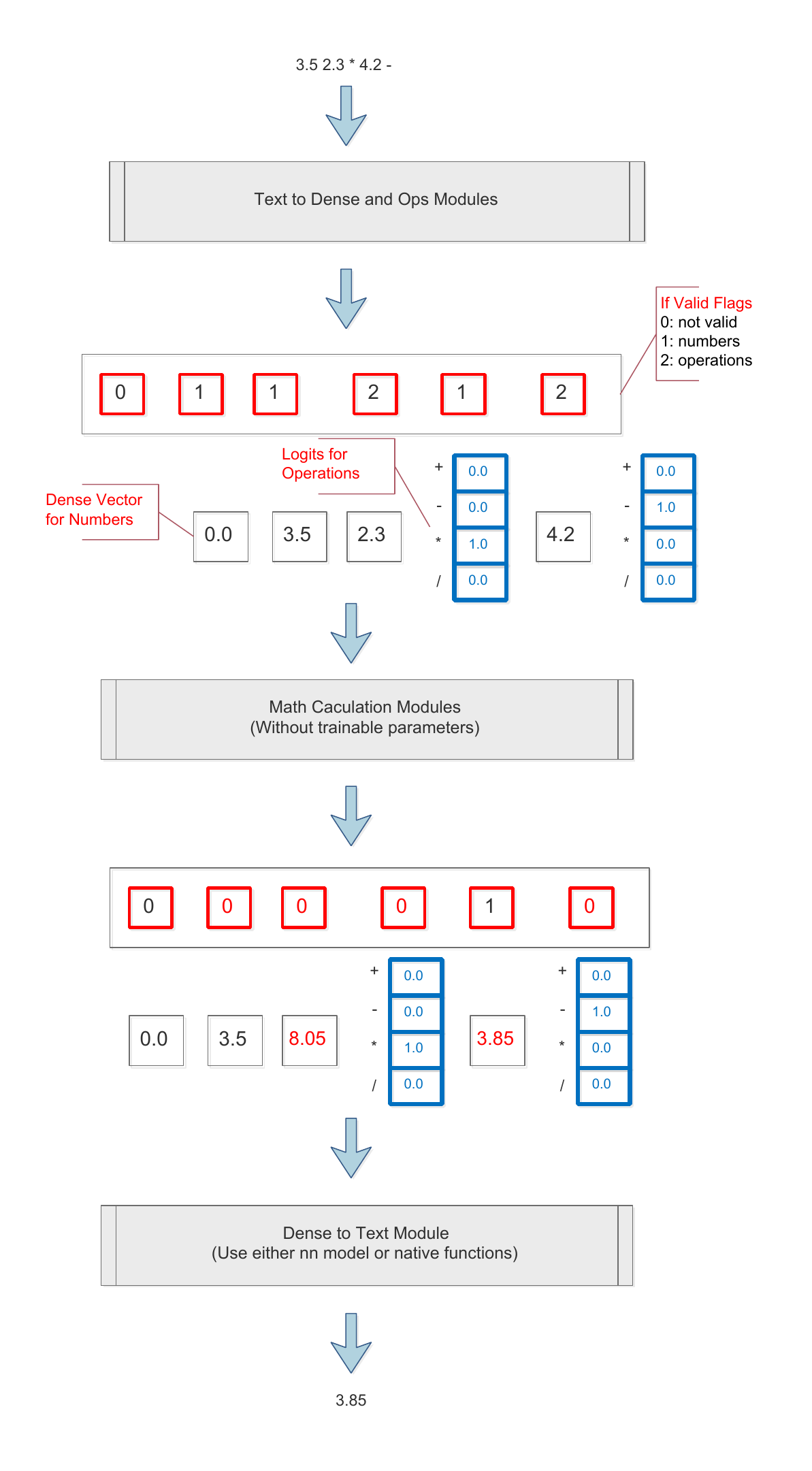}
    \caption{Overview of ArthModel. The dense number and math op conversion submodel, the arithmetic calculation submodel, and the number-to-text conversion submodel.}
	\label{fig:fig2}
\end{figure}
As shown in Figure \ref{fig:fig2}, the ArthModel can be divided into three essential parts: the dense number  and math op conversion submodel, the arithmetic calculation submodel, and the number-to-text conversion submodel.

\begin{figure}
    \centering
    \includegraphics[width=0.9\columnwidth]{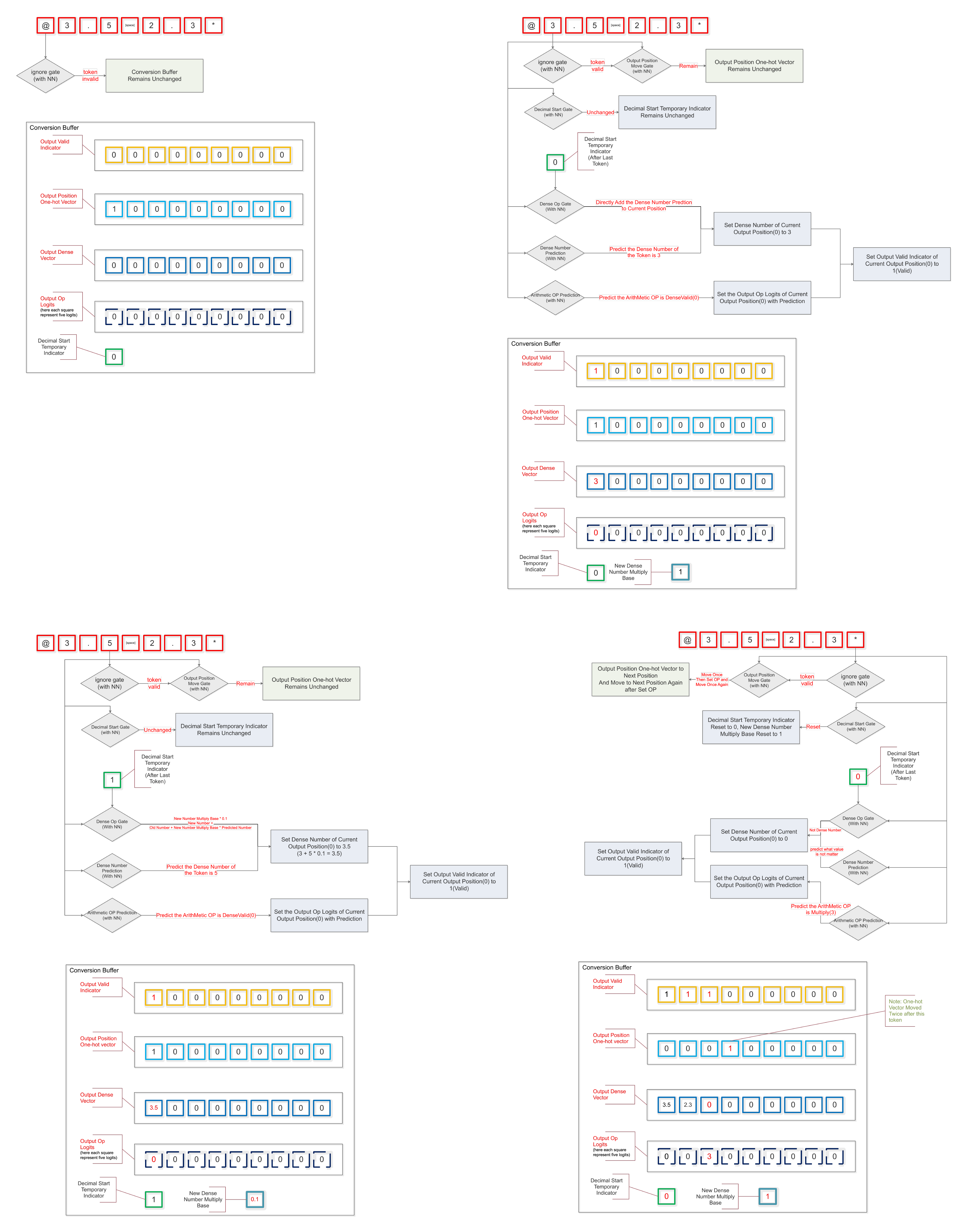}
    \caption{The dense number and math op conversion submodel. The submodel has six internal status vectors, while three of them are the outputs of the submodel, named output valid indicator, output dense vector, and output op logits. The output valid indicator vector has the same length as input tokens as well as output tokens, which the element is one when that index is dense number valid or arithmetic ops valid. The output dense vector stores the converted dense number for calculation, while the output operator(op) vector stores the predicted arithmetic ops. The output position one-hot vector indicates which output position is being considered and ensures the other position's output values remain unchanged during each conversion loop. Decimal start temporary indicator and new dense number multiply base are used to ensure the dense number conversion is accurate. }
	\label{fig:fig3}
\end{figure}

The dense number and math op conversion submodel has fewer parameters than LLM. It is a modified RNN-like module and loops for as many times as the input length. The model structure and example process is shown in Figure \ref{fig:fig3}. In the dense number and math op conversion submodel, the output position one-hot vector is initially set to zero except for the first dimension. During each loop, there is first an ignore gate for token validation, as ArthModel only process fewer arithmetic problem-related tokens such as numbers, dot, and four math op characters ('+', '-', '*', '/'). When a token is ignored, all internal status vector remains unchanged in the loop. Then, two crucial gates are predicted according to the token: the output position move gate and the decimal start gate. The output position move gate distinguishes whether the token is a math op character or part of a dense number. The decimal start gate captures a valid dot token and changes the decimal start temporary indicator when the decimal part of a dense number begins. Then, taking the token's embedding with the decimal start temporary indicator, the dense op gate predicts how to deal with the dense number in four ways, including ignorance, directly adding, adding with ten times the ordinate number, and adding after multiplying with an extra base which is updated according to the decimal start temporary indicator. Using the current token, the dense number prediction network and arithmetic operation prediction network predict the dense number and the math operation, respectively. The result of all the above predictions will also set the output valid indicator at the considered output position.

\begin{figure}
    \centering
    \includegraphics[width=0.9\columnwidth]{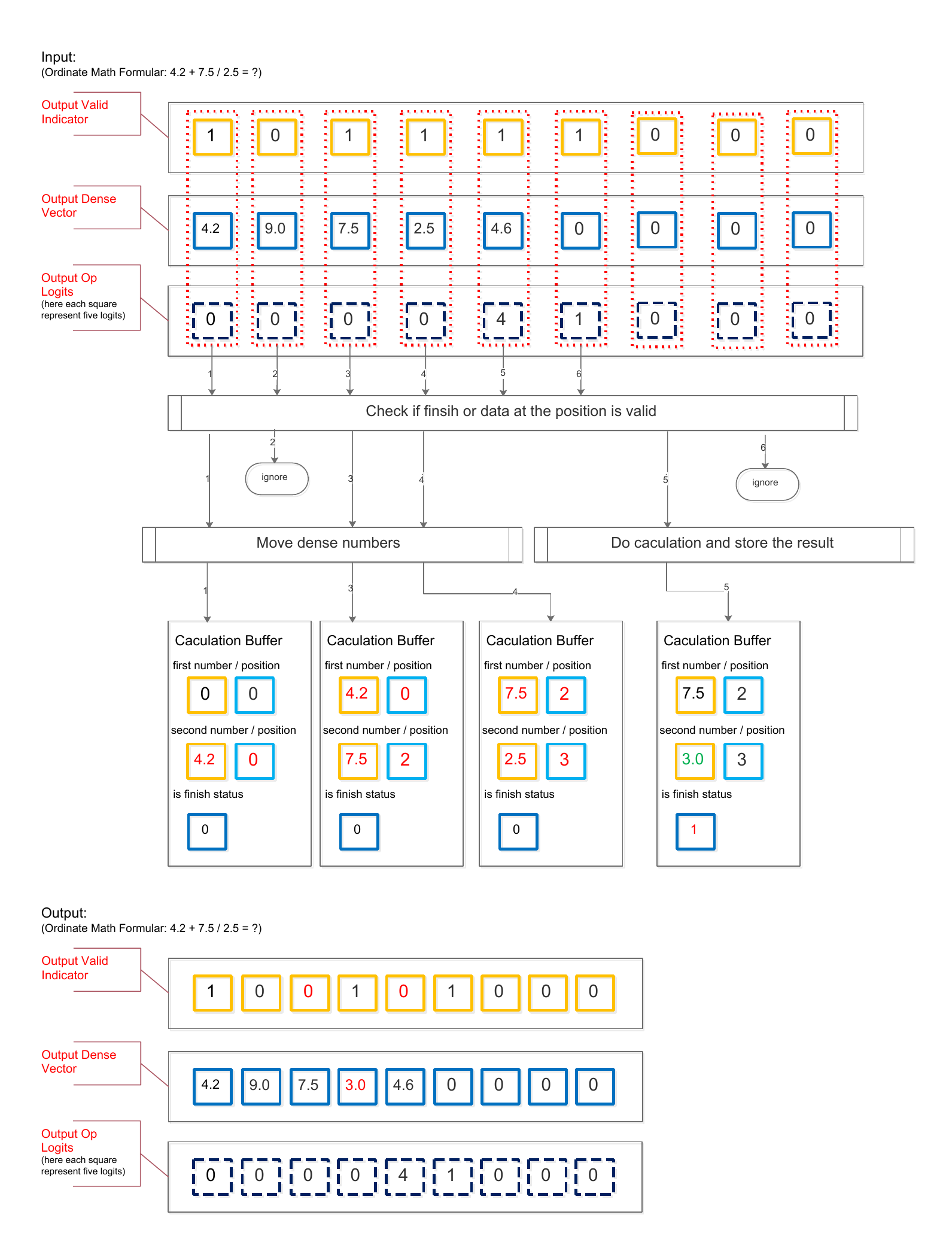}
    \caption{The progress of a single caculation module. The main idea is that record the last two dense numbers before meeting the math op, and do the math according to the binary operator.}
	\label{fig:fig4}
\end{figure}

The math calculation modules include RNN-like structure submodules, which at each step get two dense numbers and one math operation, then reduce them to one single dense number and store it properly for the next step. Figure \ref{fig:fig4} shows one math calculation submodule, and it caches the last two dense numbers until a valid math operator occurs. Then, it calculates the new number according to the math operator. Finally, it stores the result in the place of the dense number input vector where the latest number was retrieved and updates the token valid status vector that the position of the first dense number and the math operation is set to invalid status.

The dense-to-text module can use either the model transfer, which is the way the token-to-dense module does, or simply invoke a native Python API call. In our demonstration, we use the native Python 'str()' function to simplify the implementation and show the opportunity that the model incorporates with the native function call of a DL inference platform, as many other works illustrate.


\begin{table}
\begin{tabular}{|p{0.3\columnwidth}|p{0.5\columnwidth}|p{0.2\columnwidth}|}
\hline
Dataset Name                             & Example                                                                                                                                                                                                                                                                                                             & Amount of examples in the Datasets \\ \hline
Dot Place Datasets                       & \begin{tabular}[c]{@{}l@{}}1.1111\\ 11.111\\ 1111.1\\ 111.11\end{tabular}                                                                                                                                                                                                                                           & 100                                \\ \hline
Arithmetic QA Datasets                   & \{"instruction": "Please caculate this.", "input": "3 + 5 = ?", "output": "8", "swift\_express": "3 5 +"\}                                                                                                                                                                                                          & 72800                               \\ \hline
Alpace and Arithmetic QA Datasets in 4:6 & {[}\{"instruction": "Come up with a world record that does not already exist.", "input": "", "output": "World record for most number of people creating a continuous human bridge across a river."\}, \{"instruction": "Please caculate this.", "input": "3 + 5 = ?", "output": "8", "swift\_express": "3 5 +"\}{]} &   83200                        \\ \hline
\end{tabular}
\caption{Details of Datasets. We are not using full examples of a dataset to train the model}
\label{table:table1}
\end{table}

\subsection{LLM-Chat Model}
Follows standard LLaMA-Adapter process, except when AuxLLM decides to enable ArthModel for assistance, we will append the output of ArthModel to ordinate prompt which is a fixed length token with special charaster "\$" at the ends of output. We name this process "Prompt Injection". 

\section{Implementation and Training Datasets}
The training progress can be summarized in three main parts: the ArthModel training, postfix sequence prediction training of AuxLLM, and the joined training. For training different components, we construct distinct datasets for that, and a summary can be found in Table \ref{table:table1}.

The only trainable component of the ArthModel in our implementation is the dense number and math op conversion submodel, so training the ArthModel is indeed training the dense number and math op conversion submodel. Among all gates of the dense number and math op conversion submodel, the ignore gate and the decimal start gate are crucial. We construct a small dataset named "Dot Place Dataset", containing only one dense number for each line with different decimal places. As the tokenizer of LLM assigns each character as a token, we can easily generate the ground truth for each gate given the token. Moreover, to stabilize the prediction of gates in the ArthModel, we use the argmax function to generate the output of gate in prediction while using binary cross entropy loss at training, and we use the same trick to finetune the LLM model later. When the submodel is able to convert the right number into dense, it is then trained by a new dataset named "Numbers and Math Ops Dataset". It contains many numbers and math op characters are placed in the same line separated by white space. To better train math ops conversion, some lines contain only math ops.

To train the AuxLLM, we generate a dataset following the data structure of Alpaca, named "Arithmetic QA Dataset". The "instruction" field for the math question is fixed to "Please calculate this.". The arithmetic problem is placed into the "input" field, and the "output" field contains the right answer. An extra field called "swift\_express" is the postfix expression of the math question, which is also the output we want the AuxLLM to learn and generate. We begin with the easy dataset, which contains only a few operations, and the numbers are smaller than 100 with at most two digits of the decimal part. Then, to enhance the sense of operation priority, we add some arithmetic questions that add and minus ops are followed by multiply and division ops. Details can also be found in Table \ref{table:table1}.

The joined training starts with the pre-trained AuxLLM modules and ArthModel and takes the "Arithmetic QA Dataset" first. To stabilize the training process, we fixed the parameter of ArthModel at the beginning. We split training examples into small epochs with 50 examples in each epoch and train the same epoch five times before moving to the next epoch, which makes the model output as accurate result as possible before meeting new data. After about a thousand small epoch training,  we relax the fixed ArthModel parameters and continue for another a thousand small epoch training. Then, we generate a mixed training dataset named "Alpace and Arithmetic QA in 4:6 Dataset", which contains 60\% of arithmetic problems and 40\% of ordinate Alpace training data and use it for another a thousand small epoch training. Each epoch requires three repetitions before moving on to the next one.

\section{Interesting Topic and Discoveries}
The postfix expression generating is not tested on natural languages or natural arithmetic questions such as 'add three to five' or 'give you three bananas on day one and five bananas on day two, how many bananas do you get in those two days?'. Nevertheless, we think it should be possible to do so.

When we trains the model before using 'the Alpace and Arithmetic QA in 4:6 Dataset', the model can also generate readable answers for natural problems such as "Design a logo for a food store." We find it will use a logical form to give the answer such as "1.A 2.B 3.C".

\section{Thanks}
Thanks to my girlfriend, who gave me more time and cheered me up to finish this work.

\bibliographystyle{unsrtnat}
\bibliography{references}  






\end{document}